\documentclass[10pt,a4paper]{article}
\usepackage[utf8]{inputenc}
\usepackage{amsmath}
\usepackage{amsfonts}
\usepackage{amssymb}
\usepackage{graphicx}
\usepackage{booktabs}
\usepackage{array}
\usepackage[margin=1in]{geometry}
\usepackage{xcolor}
\usepackage{tcolorbox}
\usepackage{amsthm}
\usepackage{tikz}
\usetikzlibrary{arrows.meta,positioning}
\usepackage{multirow}
\newtheorem{theorem}{Theorem}
\usepackage{cite} 

\definecolor{goodcolor}{RGB}{0,150,0}
\definecolor{badcolor}{RGB}{200,0,0}
\definecolor{neutralcolor}{RGB}{100,100,100}

\newcommand{\E}{\mathbb{E}}
\newcommand{\grad}{\nabla}
\newcommand{\probpi}{\pi_\theta}
\newcommand{\logit}{z}
\newcommand{\param}{\theta}
\newcommand{\defineas}{\triangleq}
\newcommand{\goodupdate}[1]{\textcolor{goodcolor}{#1}}
\newcommand{\badupdate}[1]{\textcolor{badcolor}{#1}}

\newtcolorbox{keyinsight}{
    colback=blue!5!white,
    colframe=blue!75!black,
    title=Key Insight,
    fonttitle=\bfseries
}

\newtcolorbox{fundamentalbox}{
    colback=yellow!10!white,
    colframe=orange!75!black,
    fonttitle=\bfseries
}

\title{\textbf{Logit Dynamics in Softmax Policy Gradient Methods}}
\author{Yingru Li\thanks{Email: \texttt{szrlee@gmail.com}}}
\date{June 13, 2025}

\begin{document}

\maketitle

\begin{abstract}
We analyzes the logit dynamics of softmax policy gradient methods. We derive the exact formula for the L2 norm of the logit update vector:
$$
\|\Delta \mathbf{z}\|_2 \propto \sqrt{1-2P_c + C(P)}
$$
This equation demonstrates that update magnitudes are determined by the chosen action's probability ($P_c$) and the policy's collision probability ($C(P)$), a measure of concentration inversely related to entropy. Our analysis reveals an inherent self-regulation mechanism where learning vigor is automatically modulated by policy confidence, providing a foundational insight into the stability and convergence of these methods.
\end{abstract}

\section{Introduction}

Policy gradient methods~\cite{sutton1999policy,agarwal2021theory} update parameters based on the product of three factors: learning rate, advantage estimate, and score function. In softmax policies, the score function has a specific structure that creates rich, self-regulating dynamics. This analysis dissects these dynamics by focusing on the logits—the raw outputs of a model before the softmax normalization.

By focusing on the dynamics of the logits themselves, we can gain fundamental insights into the learning process that are independent of the specific policy parameterization $\theta$ (e.g., whether the policy is a linear model or a deep neural network). This analysis reveals the intended change in the policy's output distribution, offering a clear view of the adaptive mechanisms inherent in the softmax function.

\section{Mathematical Foundation}

This section outlines the core mathematical concepts, clarifying how the general policy gradient update translates into specific, structured updates for the policy's logits.

\subsection{Policy Gradient Framework}

The policy gradient theorem~\cite{sutton1999policy} provides the gradient of the expected return $J(\param)$ with respect to policy parameters $\param$:
\begin{equation}
\nabla_\param J(\param) = \mathbb{E}_{s,a} \left[ \nabla_\param \log \probpi(a|s; \param) A(s,a) \right]
\end{equation}
Here, $A(s,a)$ is the advantage function, and $\probpi(a|s; \param)$ is the policy.

For softmax policies, the probability of selecting action $a_i$ given state $s$ is determined by logits $\logit_k(s, \param)$:
\begin{equation}
\probpi(a_i|s; \param) = \frac{e^{\logit_i(s, \param)}}{\sum_{k} e^{\logit_k(s, \param)}} \defineas P_i
\end{equation}
The term $P_i$ serves as a shorthand for this probability.

\subsection{From Parameter Gradients to Logit Updates}

The update to the parameters $\param$ is driven by the gradient $\nabla_\param \log \probpi$. Using the chain rule, we can decompose this gradient's effect on the logits $\mathbf{\logit}$:
\begin{equation}
\nabla_\param \log \probpi = \left( \frac{\partial \mathbf{\logit}}{\partial \param} \right) \left( \nabla_\mathbf{\logit} \log \probpi \right)
\end{equation}
The term $\frac{\partial \mathbf{\logit}}{\partial \param}$ is the Jacobian of the logits with respect to the parameters, which depends on the specific architecture of the policy network. The second term, $\nabla_\mathbf{\logit} \log \probpi$, captures the desired change in the logits to increase the probability of an action.

This paper analyzes the dynamics of this second term. We define the \emph{logit update}, $\Delta \logit_j$, as the effective change to the logit $z_j$ that the algorithm seeks to induce for a given experience (state $s$, chosen action $a_c$, advantage $A$), scaled by a learning rate $\eta$. This provides a parameterization-agnostic view of the learning update.

\subsection{The Logit Score Function and Update Equations}

The gradient of the log-policy with respect to the logits, which we can call the logit score function, takes a remarkably simple form:

\begin{fundamentalbox}
\textbf{Fundamental Score Function for Softmax Logits}
\begin{equation}
\frac{\partial \log \probpi(a_i|s; \param)}{\partial \logit_j(s, \param)} = \delta_{ij} - P_j
\label{eq:score}
\end{equation}
where $\delta_{ij}$ is the Kronecker delta (1 if $i=j$, 0 otherwise), and $P_j = \probpi(a_j|s; \param)$.
\end{fundamentalbox}

The logit update, $\Delta \logit_j(s, \param)$, follows from applying the principle of stochastic gradient ascent to the objective. For a single experience $(s, a_c, A)$, the update is:
\begin{equation}
\Delta \logit_j(s, \param) = \eta \cdot A \cdot \left( \frac{\partial \log \probpi(a_c|s; \param)}{\partial \logit_j(s, \param)} \right)
\label{eq:logit_update_general_def}
\end{equation}
Substituting the score function from Equation \ref{eq:score} (with $i=c$) into Equation \ref{eq:logit_update_general_def} yields the detailed logit update equations:

\begin{tcolorbox}[colback=yellow!10!white,colframe=orange!75!black]
\textbf{Logit Update Equations}

For action $a_c$ chosen with advantage $A \equiv A(s,a_c)$:
\begin{align}
\Delta \logit_c(s, \param) &= \eta (1 - P_c) A &\text{(chosen action's logit)} \label{eq:update_chosen_logit} \\
\Delta \logit_o(s, \param) &= -\eta P_o A &\text{(other actions' logits, } o \neq c \text{)} \label{eq:update_other_logit}
\end{align}
\end{tcolorbox}

An important consequence of these update rules is the conservation of the sum of logit changes:

\begin{tcolorbox}[colback=gray!5!white,colframe=gray!75!black]
\textbf{Conservation Property}
The sum of logit updates is zero:
\begin{equation}
\sum_{j} \Delta \logit_j(s, \param) = \eta A \left( (1-P_c) + \sum_{o \neq c} (-P_o) \right) = \eta A \left(1 - \sum_j P_j \right) = \eta A (1 - 1) = 0
\end{equation}
This zero-sum property is significant. It shows that the updates adjust the \textit{relative} strengths of the logits without altering their sum. This ensures that changes to the probability distribution arise from a \emph{structured redistribution} of probability mass, rather than from a uniform shift across all logits (which the softmax function would render irrelevant to the final probabilities anyway).
\end{tcolorbox}

\section{Analysis of Logit Update Dynamics}

\subsection{Update Dynamics for Chosen Action}

The factor $(1-P_c)$ creates a natural modulation of update strength, as described in Table~\ref{tab:chosen_dynamics}:

\begin{table}[htbp]
\centering
\caption{Logit Dynamics for Chosen Action}
\label{tab:chosen_dynamics}
\begin{tabular}{@{}cccccc@{}}
\toprule
\textbf{$A$ Sign} & \textbf{$P_c$} & \textbf{Update Scaler} & \textbf{Update} & \textbf{Magnitude} & \textbf{Dynamics} \\
\midrule
\multirow{3}{*}{\goodupdate{$A > 0$}} 
& 0.1 & 0.9 & \goodupdate{$\uparrow\uparrow\uparrow$} & Large & Rapid increase \\
& 0.5 & 0.5 & \goodupdate{$\uparrow\uparrow$} & Moderate & Steady increase \\
& 0.9 & 0.1 & \goodupdate{$\uparrow$} & Small & Slow increase \\
\midrule
\multirow{3}{*}{\badupdate{$A < 0$}} 
& 0.1 & 0.9 & \badupdate{$\downarrow\downarrow\downarrow$} & Large & Rapid decrease \\
& 0.5 & 0.5 & \badupdate{$\downarrow\downarrow$} & Moderate & Steady decrease \\
& 0.9 & 0.1 & \badupdate{$\downarrow$} & Small & Slow decrease \\
\bottomrule
\end{tabular}
\end{table}

\subsection{Update Dynamics for Other Actions}

Other actions receive updates proportional to their current probabilities, as described in Table~\ref{tab:other_dynamics}:

\begin{table}[htbp]
\centering
\caption{Logit Dynamics for Other Actions}
\label{tab:other_dynamics}
\begin{tabular}{@{}cccccc@{}}
\toprule
\textbf{$A$ Sign} & \textbf{$P_o$} & \textbf{Update Scaler} & \textbf{Update} & \textbf{Magnitude} & \textbf{Effect} \\
\midrule
\multirow{3}{*}{\goodupdate{$A > 0$}} 
& 0.1 & -0.1 & \badupdate{$\downarrow$} & Small & Minor reduction \\
& 0.5 & -0.5 & \badupdate{$\downarrow\downarrow$} & Moderate & Clear reduction \\
& 0.8 & -0.8 & \badupdate{$\downarrow\downarrow\downarrow$} & Large & Major reduction \\
\midrule
\multirow{3}{*}{\badupdate{$A < 0$}} 
& 0.1 & -0.1 & \goodupdate{$\uparrow$} & Small & Minor increase \\
& 0.5 & -0.5 & \goodupdate{$\uparrow\uparrow$} & Moderate & Clear increase \\
& 0.8 & -0.8 & \goodupdate{$\uparrow\uparrow\uparrow$} & Large & Major increase \\
\bottomrule
\end{tabular}
\end{table}

\section{Update Magnitude and Collision Probability}

\subsection{The Update Magnitude Formula}

The total magnitude of the logit update vector for an experience is:
\begin{equation}
\|\Delta \logit\|_2 = \sqrt{\sum_j (\Delta \logit_j)^2} = \sqrt{(\eta(1-P_c)A)^2 + \sum_{o \neq c} (-\eta P_o A)^2} = \eta |A| \sqrt{(1-P_c)^2 + \sum_{o \neq c} P_o^2}
\end{equation}

By expanding $(1-P_c)^2 = 1 - 2P_c + P_c^2$ and noting that $P_c^2 + \sum_{o \neq c} P_o^2 = \sum_a P_a^2$, this can be rewritten in a more revealing form:

\begin{fundamentalbox}
\textbf{Logit Update Magnitude Formula}
\begin{equation}
\|\Delta \logit\|_2 = \eta |A| \sqrt{1 - 2P_c + C(P)}
\label{eq:update_magnitude}
\end{equation}
where $C(P) = \sum_a P_a^2$ is the collision probability.
\end{fundamentalbox}

\subsection{Understanding Collision Probability}

The collision probability $C(P)$ is a fundamental measure of distribution concentration:

\begin{keyinsight}
\textbf{Collision Probability Properties:}
\begin{itemize}
\item \textbf{Definition}: $C(P) = \sum_a P_a^2$ - the probability that two independent samples from $P$ are identical
\item \textbf{Range}: $1/n \leq C(P) \leq 1$ for $n$ actions
\item \textbf{Minimum}: $C(P) = 1/n$ when $P$ is uniform (maximum entropy/spread)
\item \textbf{Maximum}: $C(P) = 1$ when $P$ is deterministic (minimum entropy/concentration)
\item \textbf{Information theoretic}: $C(P) = e^{-H_2(P)}$ where $H_2$ is the Rényi entropy of order 2
\end{itemize}
\end{keyinsight}

\subsection{How Collision Probability Shapes Learning}

The logit update magnitude formula reveals how $C(P)$ controls learning dynamics:

\begin{equation}
\|\Delta \logit\|_2 = \eta |A| \underbrace{\sqrt{1 - 2P_c + C(P)}}_{\text{sensitivity factor}}
\end{equation}

This sensitivity factor depends on two components:
\begin{enumerate}
\item \textbf{Action-specific term}: $(1 - 2P_c)$ - larger when choosing an unlikely action ($P_c \to 0$).
\item \textbf{Distribution-wide term}: $C(P)$ - larger for more concentrated (low-entropy) distributions.
\end{enumerate}

\begin{tcolorbox}[colback=blue!5!white,colframe=blue!75!black]
\textbf{Key Dynamics:}
\begin{itemize}
\item When $P_c \approx 0$ (exploring): $\|\Delta \logit\|_2 \approx \eta |A| \sqrt{1 + C(P)}$. The update is large, especially if the policy is already concentrated elsewhere (high $C(P)$).
\item When $P_c \approx 1$ (exploiting): $\|\Delta \logit\|_2 \approx \eta |A| \sqrt{1 - 2 + 1} = 0$, since $P_c \approx 1$ implies $C(P) \approx 1$. Confident choices lead to minimal adjustments.
\item Maximum possible: $\|\Delta \logit\|_2 \approx \eta |A| \sqrt{2}$ when $P_c \to 0$ and another action is deterministic, so $C(P) \to 1$.
\end{itemize}
\end{tcolorbox}

\section{Conclusion}

The mathematical structure of softmax policy gradients, when analyzed through its effect on logits, reveals an elegant self-regulating system for learning. Our detailed analysis of these logit dynamics highlights several key aspects:

\begin{enumerate}
    \item \textbf{Probability-Dependent Individual Logit Updates:} The change to the chosen action's logit ($\Delta \logit_c$) is scaled by $(1-P_c)$, ensuring that rarely-chosen actions (low $P_c$) receive proportionally larger updates, facilitating rapid learning from surprising experiences. Conversely, highly probable actions ($P_c \approx 1$) see minimal logit changes, promoting stability. Updates to other logits ($\Delta \logit_o$) are scaled by their own probabilities $P_o$, ensuring a coherent redistribution of probability mass.

    \item \textbf{Conservation of Relative Logit Strength:} The property $\sum_j \Delta \logit_j = 0$ is a cornerstone of the logit dynamics. It guarantees that updates are a zero-sum game, redistributing influence among actions rather than shifting all logits up or down uniformly. This is the mechanism that forces a structured change in the resulting probability distribution.

    \item \textbf{Update Magnitude Governed by Chosen Action and Distribution Concentration:} The L2 norm of the entire logit update vector, $\|\Delta \logit\|_2 = \eta |A| \sqrt{1 - 2P_c + C(P)}$, reveals a direct link to the probability of the chosen action $P_c$ and the collision probability $C(P) = \sum_a P_a^2$.
        \begin{itemize}
            \item $C(P)$ is a measure of distribution concentration, inversely related to entropy (e.g., $C(P)=e^{-H_2(P)}$). Low entropy implies high $C(P)$.
            \item When exploring (chosen action $P_c \approx 0$), the update magnitude becomes large: $\|\Delta \logit\|_2 \approx \eta |A| \sqrt{1 + C(P)}$. This reflects a significant attempt to shift the policy.
            \item When exploiting a dominant action ($P_c \approx 1$, so $C(P) \approx 1$), the update magnitude vanishes: $\|\Delta \logit\|_2 \approx 0$. This shows that confident choices lead to minimal overall changes in the logit vector, promoting convergence.
        \end{itemize}
    This shows that the overall "vigor" of the logit updates is modulated by both the specific action chosen ($P_c$) and the overall concentration ($C(P)$) of the policy, which is intrinsically linked to policy entropy.

    \item \textbf{Natural Learning Trajectory:} The interplay between $P_c$ and $C(P)$ facilitates a natural transition from exploration to exploitation. Initially, when the policy is uncertain (high entropy, low $C(P)$), it can explore and adapt readily. As it learns and becomes confident (low entropy, high $C(P)$), the updates for those confident actions diminish, leading to stability.
\end{enumerate}

These properties emerge directly from the mathematical definition of the softmax function and the policy gradient theorem. Understanding these precise logit dynamics, especially the role of collision probability in modulating the overall update strength, provides crucial insights for designing algorithms, tuning hyperparameters, and diagnosing issues like premature convergence or insufficient exploration.

\bibliographystyle{plain} 
\bibliography{references}  

\begin{thebibliography}{1}

\bibitem{agarwal2021theory}
Alekh Agarwal, Sham~M Kakade, Jason~D Lee, and Gaurav Mahajan.
\newblock On the theory of policy gradient methods: Optimality, approximation, and distribution shift.
\newblock {\em Journal of Machine Learning Research}, 22(98):1--76, 2021.

\bibitem{sutton1999policy}
Richard~S Sutton, David McAllester, Satinder Singh, and Yishay Mansour.
\newblock Policy gradient methods for reinforcement learning with function approximation.
\newblock {\em Advances in neural information processing systems}, 12, 1999.

\end{thebibliography}

\appendix

\section{Collision Probability, Rényi Entropy, and Shannon Entropy}
\label{sec:collision_entropy}

\subsection{Definitions and Relationships}
\label{ssec:entropy_definitions}

Understanding the exploration-exploitation characteristics of a policy often involves measuring the uncertainty or concentration of its action probability distribution. Several related concepts are key:

For a discrete probability distribution $P = (p_1, \ldots, p_n)$:

\begin{itemize}
    \item \textbf{Collision Probability ($C(P)$):} This is the probability that two independent samples drawn from the distribution $P$ will be the same.
    \begin{equation}
        C(P) = \sum_{i=1}^n p_i^2
    \end{equation}
    \item \textbf{Rényi Entropy of order $\alpha$ ($H_\alpha(P)$):} This is a family of entropy measures. For $\alpha > 0, \alpha \neq 1$:
    \begin{equation}
        H_\alpha(P) = \frac{1}{1-\alpha} \log \sum_{i=1}^n p_i^\alpha
    \end{equation}
    (We assume base $e$ for the logarithm, for consistency with $\log$ meaning natural logarithm.)
    \item \textbf{Rényi Entropy of order 2 ($H_2(P)$):} A special case of Rényi entropy that directly relates to the collision probability:
    \begin{equation}
        H_2(P) = \frac{1}{1-2} \log \sum_{i=1}^n p_i^2 = -\log \sum_{i=1}^n p_i^2 = -\log C(P)
    \end{equation}
    From this, it's clear that the collision probability can be expressed as:
    \begin{equation}
        C(P) = e^{-H_2(P)}
    \end{equation}

    \item \textbf{Shannon Entropy ($H(P)$):} This is the standard measure of uncertainty in information theory. It is also the limit of the Rényi entropy as $\alpha \to 1$:
    \begin{equation}
        H(P) = \lim_{\alpha \to 1} H_\alpha(P) = -\sum_{i=1}^n p_i \log p_i
    \end{equation}
\end{itemize}

Derivation of the Shannon entropy as the limit of Rényi entropy:
The limit $\lim_{\alpha \to 1} \frac{\log \sum p_i^\alpha}{1-\alpha}$ is of the form $\frac{0}{0}$. Applying L'Hôpital's Rule:
\begin{align*}
    \lim_{\alpha \to 1} \frac{\frac{d}{d\alpha} \left( \log \sum_i p_i^\alpha \right)}{\frac{d}{d\alpha} (1-\alpha)} &= \lim_{\alpha \to 1} \frac{\frac{1}{\sum_i p_i^\alpha} \sum_i (p_i^\alpha \log p_i)}{-1} \\
    &= \frac{\frac{1}{\sum_i p_i^1} \sum_i (p_i^1 \log p_i)}{-1} \\
    &= \frac{\sum_i p_i \log p_i}{-1} = -\sum_i p_i \log p_i
\end{align*}

\subsection{Qualitative Relationships and the Entropy Ordering Theorem}
\label{ssec:entropy_ordering}

While individual action probabilities $p_i$ directly influence logit update magnitudes, the collision probability $C(P)$ as a summary statistic provides insight into the distribution's concentration. It relates to Shannon entropy via the following theorem:

\begin{theorem}[Entropy Ordering]
\label{thm:entropy_ordering}
For any probability distribution $P$:
\begin{equation}
    H_2(P) \leq H(P)
\end{equation}
This implies:
\begin{equation}
    C(P) \geq e^{-H(P)}
\end{equation}
Equality holds in both inequalities if and only if $P$ is uniform over its support.
\end{theorem}

\begin{proof}
The definition of Rényi entropy of order 2 is $H_2(P) = -\log C(P) \quad (*)$. A fundamental property of Rényi entropies is that $H_\alpha(P)$ is a non-increasing function of $\alpha$. That is, for $0 < \alpha < \beta$, $H_\alpha(P) \geq H_\beta(P)$. Since Shannon entropy $H(P)$ is the limit as $\alpha \to 1$, we can consider it $H_1(P)$. As $1 < 2$, it follows that $H_1(P) \geq H_2(P)$, so $H(P) \geq H_2(P)$. Substituting equation $(*)$ into this inequality:
$H(P) \geq -\log C(P)$.
Multiplying by -1 reverses the inequality: $-H(P) \leq \log C(P)$. Exponentiating both sides preserves the inequality: $e^{-H(P)} \leq C(P)$.
\end{proof}

This theorem establishes that $H_2(P)$ provides a lower bound on $H(P)$, and conversely, $e^{-H(P)}$ provides a lower bound on $C(P)$.

\subsection{Comparative Behavior}
\label{ssec:entropy_comparison}

The different entropy measures capture the "peakedness" or "flatness" of a distribution.

\begin{table}[h!]
\centering
\small
\caption{Entropy Measures for Common Distributions (using natural logarithm)}
\label{tab:entropy_measures_revised}
\begin{tabular}{@{} l c c c @{}}
\toprule
\textbf{Distribution Type} & \textbf{$C(P)$} & \textbf{$H_2(P) = -\log C(P)$} & \textbf{$H(P)$} \\
\midrule
Uniform on $n$ actions ($p_i = 1/n$) & $1/n$ & $\log n$ & $\log n$ \\
\rule{0pt}{4ex}
Two-point ($[p, 1-p]$) & $p^2 + (1-p)^2$ & $-\log(p^2 + (1-p)^2)$ & $-p\log p - (1-p)\log(1-p)$ \\
\rule{0pt}{4ex}
Near-deterministic ($p_k \approx 1$) & $\approx 1$ & $\approx 0$ & Small positive $(\approx 0)$ \\
\bottomrule
\end{tabular}
\end{table}

\textbf{Key Insights:}
\begin{itemize}
    \item \textbf{Relationship to Logit Dynamics:} The individual probabilities $P_i$ directly appear in the logit update rules. The collision probability $C(P) = \sum P_i^2$ summarizes the overall concentration of these probabilities. A high $C(P)$ (low entropy) often implies that some $P_i$ are large, which can lead to smaller updates for those actions if chosen (due to the $1-P_c$ term) and larger negative updates for other actions if their $P_o$ is still significant.
    \item \textbf{Ordering of Distributions:} All these measures agree on the ordering of distributions from most uniform (spread out) to most deterministic (concentrated).
    \begin{itemize}
        \item Uniform: $C(P)$ is minimal ($1/n$), $H_2(P)$ and $H(P)$ are maximal ($\log n$).
        \item Deterministic: $C(P)$ is maximal (1), $H_2(P)$ and $H(P)$ are minimal (0).
    \end{itemize}
    \item \textbf{Computational Simplicity:} $C(P)$ is computationally simpler than $H(P)$ as it avoids logarithms.
    \item \textbf{Interpretive Value:} $C(P)$ and $H_2(P)$ offer a direct perspective on the "self-collision" likelihood or squared concentration of the policy, which this paper shows is fundamental to the magnitude of the logit update vector. $H(P)$ provides the well-established information-theoretic interpretation of average uncertainty.
\end{itemize}

\end{document}